# Cherry Yield Forecast: Harvest Prediction for Individual Sweet Cherry Trees

**Andreas Gilson[1], Peter Pietrzyk[1], Chiara Paglia[1], Annika Killer[2], Fabian Keil[1], Lukas Meyer[3], Dominikus Kittemann[2], Patrick Noack[2] and Oliver Scholz[1]**

[1] Fraunhofer Institute for Integrated Circuits IIS (IIS/EZRT), Fürth, Germany
[2] University of Applied Sciences Weihenstephan-Triesdorf, Merkendorf, Germany
[3] Friedrich-Alexander-University Erlangen-Nürnberg, Erlangen, Germany

***Abstract.***

*The digitization of orchards is an ongoing research topic, with the concept of digital twins attracting growing interest in recent times. Obtaining objective information from dynamic natural scenes is often challenging but has the potential to open the doors for a smarter kind of agriculture with various benefits for involved stakeholders. This paper is part of a publication series from the For5G project that has the goal of creating digital twins of sweet cherry trees. At the beginning a brief overview of the previous work in this project is provided, as well as a detailed review of literature closely related to this topic.* Afterwards the focus shifts to a crucial problem in the fruit farming domain: the difficulty of making reliable yield predictions early in the season. Following three Satin sweet cherry trees along the year 2023 enabled the collection of accurate ground truth data about the development of cherries from dormancy until harvest. The methodology used to collect this data is presented, along with its evaluation and visualization. The predictive power of counting objects at all relevant vegetative stages of the fruit development cycle in cherry trees with regards to yield predictions is investigated. It is found that all investigated fruit states are suitable for yield predictions based on linear regression. Conceptionally, there is a trade-off between earliness and external events with the potential to invalidate the prediction. Considering this, two optimal timepoints are suggested that are opening cluster stage before the start of the flowering and the early fruit stage right after the second fruit drop. However, both timepoints are challenging to solve with automated procedures based on image data. Counting developing cherries based on images is exceptionally difficult due to the small fruit size and their tendency to be occluded by leaves. It was not possible to obtain satisfying results relying on a state-of-the-art fruit-counting method. Counting the elements within a bursting bud is also challenging, even when using high resolution cameras. It is concluded that accurate yield prediction for sweet cherry trees is possible when objects are manually counted and that automated features extraction with similar accuracy remains an open problem yet to be solved.

## Introduction

The continuous development and integration of new technologies have been transformative across various sectors, including agriculture, with notable advancements seen within the field of horticulture (Kondratieva et al., 2022). Smart farming and precision agriculture share a fundamental dependence on precise and reliable real-world data. One possible way to address this is by using digital twins (DT). The main function of a DT is to capture the complexities of a living organism and its interactions with the surrounding environment (Pylianidis et al., 2021). The significance of DTs in academic and applied fields is growing, underscoring their pivotal role in future agricultural strategies and robotic applications (Nasirahmadi & Hensel, 2022; Purcell & Neubauer, 2023). This paper presents the ongoing work regarding yield prediction within the *For5G* project, which ultimately aims to solve the challenges of creating DTs for fruit trees. After reviewing the related literature and presenting background information of our project this paper addresses the question of whether it is possible to accurately predict the expected fruit yield of individual sweet cherry trees at harvest.

The precise contributions of this paper are as follows:

- Proposal of a method for gathering accurate data for yield prediction in orchards.

- Conceptual embedding of yield prediction into our comprehensive project framework for the creation of digital twins of fruit trees.

- Evaluation of ground truth data from three cherry trees from the start of the 2023 season until late May 2024, with insights into timepoint and accuracy of possible yield predictions.

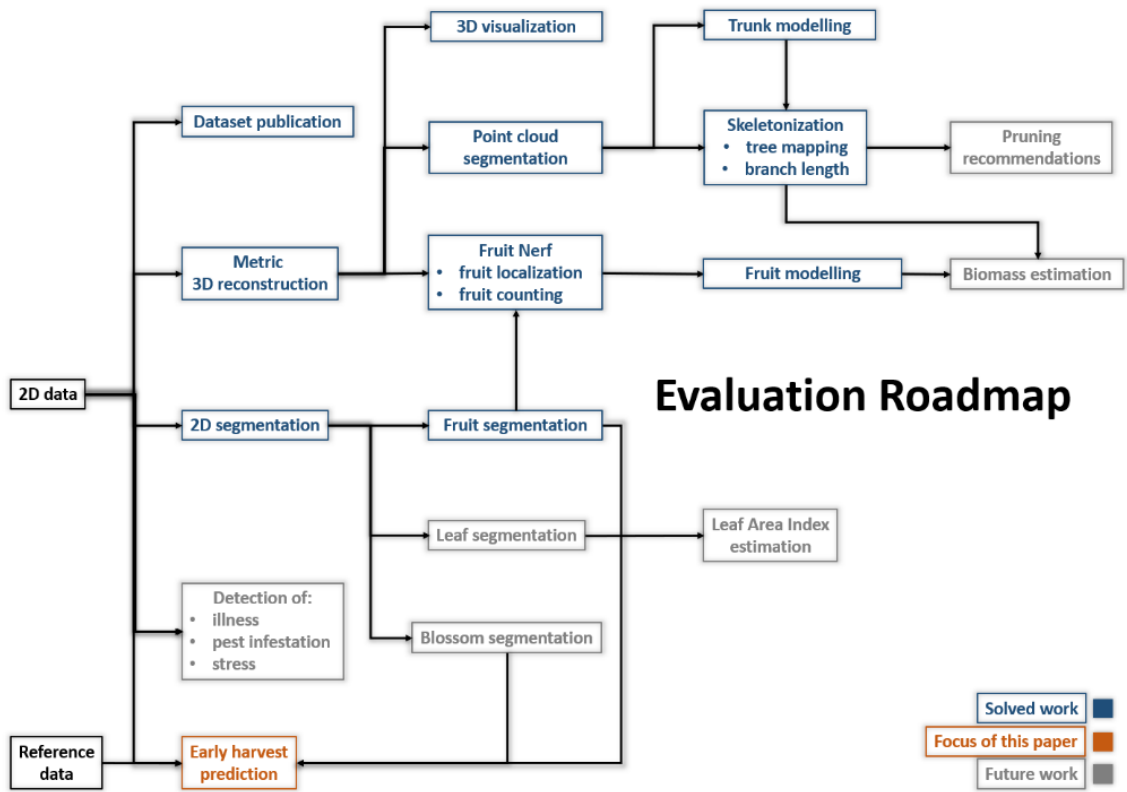

**Fig 1.** *For5G* project evaluation roadmap with distinction between already solved, future and current work.

The data used for the following analysis has been published as *CherrySet* (Gilson et al., 2023), and is openly available to download free of charge[1]. The project is accompanied by several conceptual and technical publications including tree skeletonization (Meyer, Gilson, Scholz, et al., 2023), data encoding of DTs using knowledge graphs (Gilson et al., 2024) and accurate fruit counting (Meyer et al., 2024). Fig. 1 displays an overview of our evaluation roadmap and puts the work of this paper into the project context.

## Related Work

Based on research results from various fields, DTs show great potential for improving efficiency and sustainability (Slob & Hurst, 2022). The use of DTs in the field of agriculture is still at an early stage (Verdouw et al., 2021). Monteiro et al. (2018) presented a model for the introduction of DTs in sustainable agriculture. The model shows how DTs can contribute to the life cycle of planning, operation, monitoring, and optimization of agricultural products. Skobelev et al. (2020) investigate DTs in agricultural business management. The paper proposes a multi-agent approach for the development of DTs of crops that reflects the different stages of crop development and allows to predict the harvest more accurately. In addition, the possibility of formalizing domain knowledge about new agricultural technologies for crop production and the automation of decision-making processes in precision agriculture is demonstrated (Skobelev et al., 2020). In the article by Khatraty et al. (2023), an architecture for rice field management with DTs is presented that combines IoT sensors, satellite data and machine learning models to predict crop yields, weather, and soil conditions. The concept of sDTs is also growing in the greenhouse sector (Khatraty et al., 2023). Howard et al. (2020) propose the integration of DTs into greenhouse management to optimize energy consumption and increase productivity. In summary, the management aspects of using DTs to plan, monitor, control and optimize agricultural processes need to be further investigated (Verdouw et al., 2021).

Yield data is of significant importance for the fruit industry. Yield estimations allow decisions to be made about orchard management in terms of the labor required, storage, transport, and marketing. Fruit growers generally use traditional yield estimation methods, where yields are estimated by manually counting or weighing samples from small, randomly selected areas to then estimate the yield of the entire orchard or a large area (He et al., 2022). Wulfsohn et al.(2012) and Marani et al. (2021) point out that due to the differences between individual fruit trees and the variability of many orchard parameters, the error rate of such approaches is higher than expected. In addition, it is emphasized that estimating fruit yields requires extensive sampling and counting of fruit, which makes it difficult to collect pre-harvest yield data. Therefore, the introduction of an efficient automated system for modern fruit crop management is considered crucial to reduce the manual effort (He et al., 2022). Another approach for obtaining yield data is to recognize and count the fruits directly on the plants and then estimate the fruit yield. Early detection of fruit on trees is based on image processing methods or machine learning algorithm-based classifiers that recognize fruit based on color, shape and texture characteristics, as widely discussed (Liu et al., 2018; Sengupta & Lee, 2014; Xu et al., 2019). In a review paper, Gongal et al. (2015) emphasize that machine learning can provide more accurate results in fruit recognition than the traditional methods of image processing.

Many researchers have worked on methods for predicting and estimating fruit yields. An overview of the application of image processing technologies for automatic yield mapping of fruit and vegetable crops was given by Darwin et al. (2021). Anderson et al. (2021) provided an overview of yield models and image processing technologies offered for yield mapping in orchards in a new study. They concluded that most of the current research on yield assessment in orchards is dominated by image processing. However, most of the work on yield prediction that has been examined has focused on the application of a single technology or has not been specialized in fruit production. Therefore, a comprehensive analysis of yield prediction and estimation for orchards and fruit is not available (He et al., 2022).

---

[1] Data can be downloaded at: https://fordatis.fraunhofer.de/handle/fordatis/383

Recently the field of agricultural data processing and image analysis has seen the advent of deep learning (DL). An overview of DL technologies for fruit detection and yield estimation is provided by Koirala et al. (2019a, 2019b), which also include suggestions for the use of public datasets and the implementation of transfer learning. Orchard yield estimation in most studies focuses on target recognition. However, it can also be considered as generic object counting. The cropping patterns of different fruit tree species differ considerably, some are dense (e.g. apple trees or grapevines), others are patchy (e.g. mango trees). For this reason, a variety of counting strategies have been proposed for estimating fruit tree yield, including regression counting of fruit pixel density (Zaman et al., 2010), fruit counting in independent images (Payne et al., 2013), and fruit counting in composite images (Mekhalfi et al., 2020). In previous work of the For5G project *FruitNeRF* (Meyer et al., 2024), a fruit agnostic framework for fruit counting with state of the art results was developed.

## Project Background and Methods

As briefly discussed in the introduction, the For5G project aims to develop an end-to-end approach for the creation of digital tree twins. Focusing on the sweet cherry the project started with a systematic proposal of our approach that is displayed in Fig. 2 from Meyer et al. (2023).

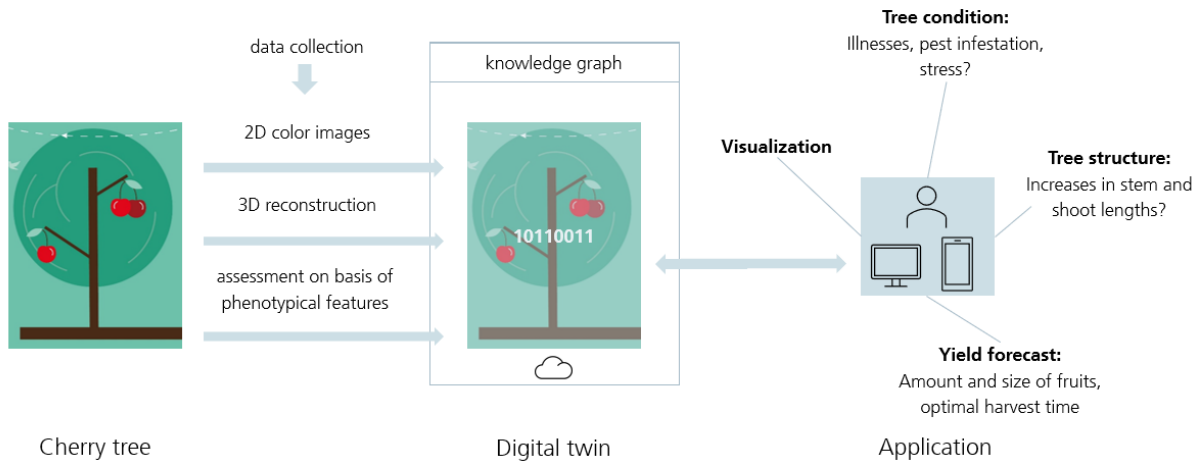

**Fig 2. Visualization of the For5G concept from a real-world tree to a polished end user application.**

### Overview of data collection process

In our data-gathering process, we utilize a system involving a drone equipped with a camera, autonomously navigating a predefined course, and transmitting data directly via 5G. Due to the absence of a native 5G network onsite, a portable campus network was utilized. The unmanned aerial vehicle (UAV) deployed for this initiative is a DJI M300, equipped with a high-resolution photogrammetric camera for capturing detailed videos and images. This configuration allows us to pre-set various flight trajectories customized to our needs, whether for close-up inspections of individual trees or broader sweeps across rows of trees. The fieldwork is conducted at the *Obstinformationszentrum Fränkische Schweiz*, a local cherry orchard research facility in Hiltpoltstein, Germany. Over the course of the 2023 season, three selected sweet cherry trees of type *Satin* were subject to detailed measurements at 12 different time points. The resulting high-resolution images and accompanying manually collected ground truth data are published in a related publication (Gilson et al., 2023). Fig. 3 gives an impression of the field site and the 5G campus network. For later evaluations, the focus was switched from detailed measurements of individual trees to row-wise data collection with less detail but also significantly shorter measurement times per tree.

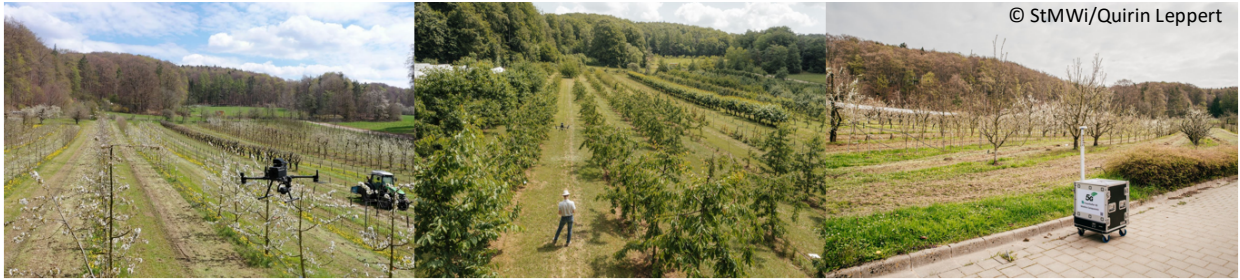

**Fig 3. Pictures of the orchard used as data site in *Hiltpoltstein*, the drone and the mobile 5G campus network station.**

## Obtaining ground truth data as reference

While the focus of the data collection was on obtaining high-quality image data, ground truth reference data was manually counted as well. To verify our automated evaluation results, real-world data is required. From their dormant state in March, through the blossoming and growth phases, up to the harvest in July 2023, the fruit development process was monitored. Counted object types include buds, blossoms, and fruit of selected branches during each vegetation phase. For this purpose, 6 branches were selected and marked on every tree to be able to track the development of individual fruit from buds to ready to harvest cherries. In Table 1 the digitized reference values for one exemplary branch are displayed. The branch was physically marked with a pink marker during the manual counts on eight different days between March and July 2023. At harvest not only the reference branches, but all the cherries of the entire tree were counted and weighted. Additionally, the distinction was made between *goodCrops* and *badCrops*, whereas *goodCrops* refer only to cherries that are in good enough condition to be sold directly to end consumers. In contrast, the term b*adCrops* stands for cherries that were damaged, showed signs of parasite infestation or were not eligible for direct sale for other reasons. However, they still could be eligible for juice production or secondary products.

**Table 1. Exemplary display of manually collected reference data for "pink" branch from tree satin_2. The fruit development was tracked from late stages of dormancy until the finally harvested cherry.**

| Date 2023 season | BBCH | treeID | branchID | branchColor | objectType | objectCount | cropWeight |
|---|---|---|---|---|---|---|---|
| Mar-2 | 51 | satin_2 | 2s1 | pink | bud | 175 | |
| Apr-14 | 56 | satin_2 | 2s1 | pink | bud | 96 | |
| Apr-25 | 60 | satin_2 | 2s1 | pink | blossom | 257 | |
| May-25 | 65 | satin_2 | 2s1 | pink | blossom | 141 | |
| Jun-06 | 75 | satin_2 | 2s1 | pink | cherry | 53 | |
| Jun-16 | 81 | satin_2 | 2s1 | pink | cherry | 52 | |
| Jul-06 | 85 | satin_2 | 2s1 | pink | cherry | 52 | |
| Jul-14 | 89 | satin_2 | 2s1 | pink | goodCrops | 31 | 0,29 |
| Jul-14 | 89 | satin_2 | 2s1 | pink | badCrops | 23 | 0,18 |
| Jul-14 | 89 | satin_2 | 2s1 | pink | totalCrops | 54 | 0,47 |

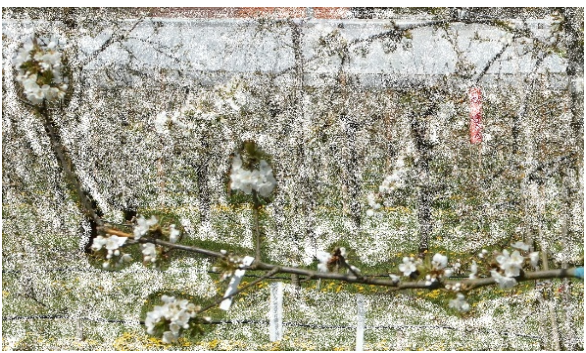 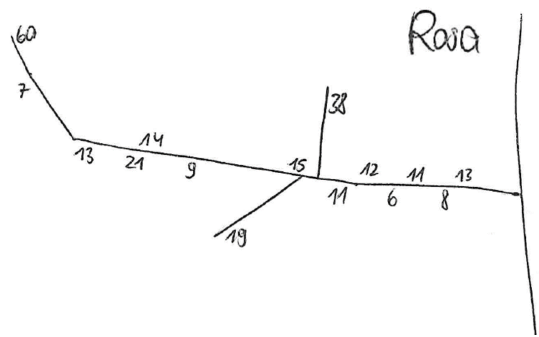

**Fig 4. Photo (left) of the "pink" branch referenced in Table 1 with noised background for better visibility. Handwritten reference data (right) for this branch on the same day (25th of April 2023).**

Fig. 4 shows a picture of the exemplary chosen "pink" branch as a picture in early flowering stage (25[th] of April 2023) together with the handwritten reference data. Not all counted blossoms are visible in the image, caused by the limited perspective given the picture is taken from one angle only.

How this exact branch can be conceptualized as a knowledge graph including the counted objects information is explained and visualized in Gilson et al. (2024). For referencing the development stages, the unified BBCH code from Meier (2018) was used. Fig. 5 from Bound et al. (2022) illustrates the early stages from dormancy up until young fruit with pictures and corresponding BBCH code for a clear categorization. For a detailed explanation of the BBCH stages and the BBCH stages of the fruit development phase refer to Meier (2018).

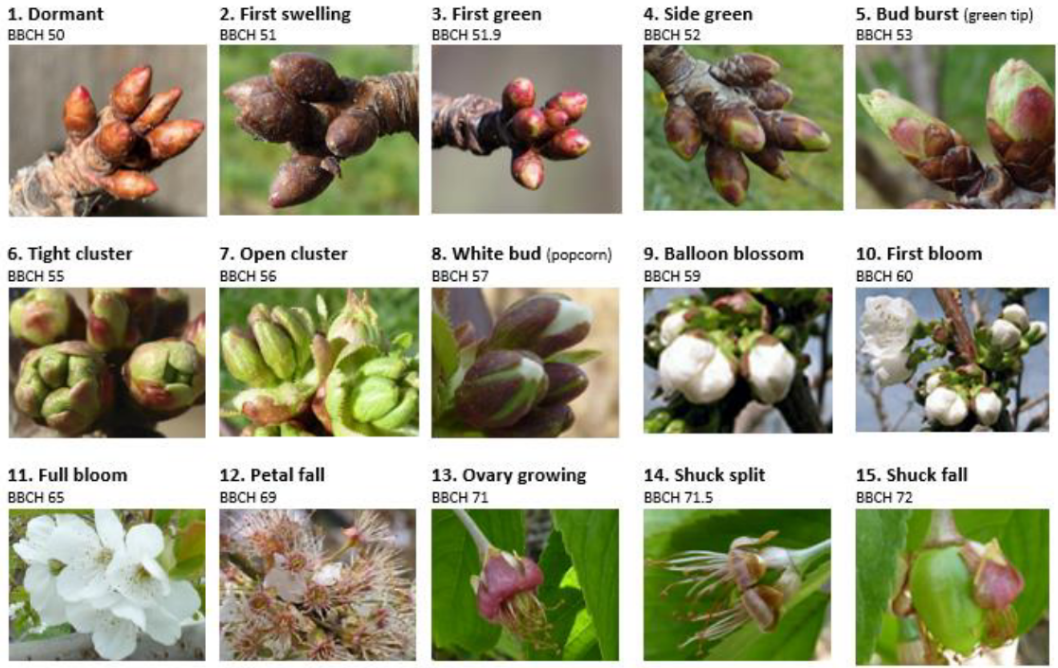

Fig 5. Visualization of the phenological stages in sweet cherry with reference to the BBCH growth stages (Meier, 2018). Photo credits: SA Bound. Figure from (Bound et al., 2022).

## Results

**Ground truth data analysis**

We followed three cherry trees over the course of the 2023 season and selected 6 branches on each of the trees for accurate ground truth sampling. In the following figures the results are visualized graphically once for all reference branches over time (Fig. 6) and aggregated by tree (Fig. 7). It can be observed that there is an initial drop in objects between early March and the middle of April. This is caused by the fact that it is not possible to distinguish buds that will become leaves from buds that result in blossoms. However, once the buds open up, they contain clusters with multiple blossoms, which explains the object increase in later April and the peak at the stage of full flowering. After that the blossom count steadily declines and young fruit start to develop. However, not every blossom produces fruit, explaining the object count decrease between April and May. This effect is reinforced during the early stages of fruit development, where trees drop a certain amount of their fruit (called "fruit drop") before they end up with the final number of cherries that could potentially make it to harvest. Cherries that are subject of the fruit drop were not counted on the 6[th] of June. Between early June and harvest during July, there is comparatively little movement in the object count. Occasionally measured increasing numbers are a result of miscounts and measurement errors, since additional cherries cannot develop naturally at this stage. The slight but steady decrease observed for the other data points can be explained by fruit that get picked by birds or are lost due to other natural influences.

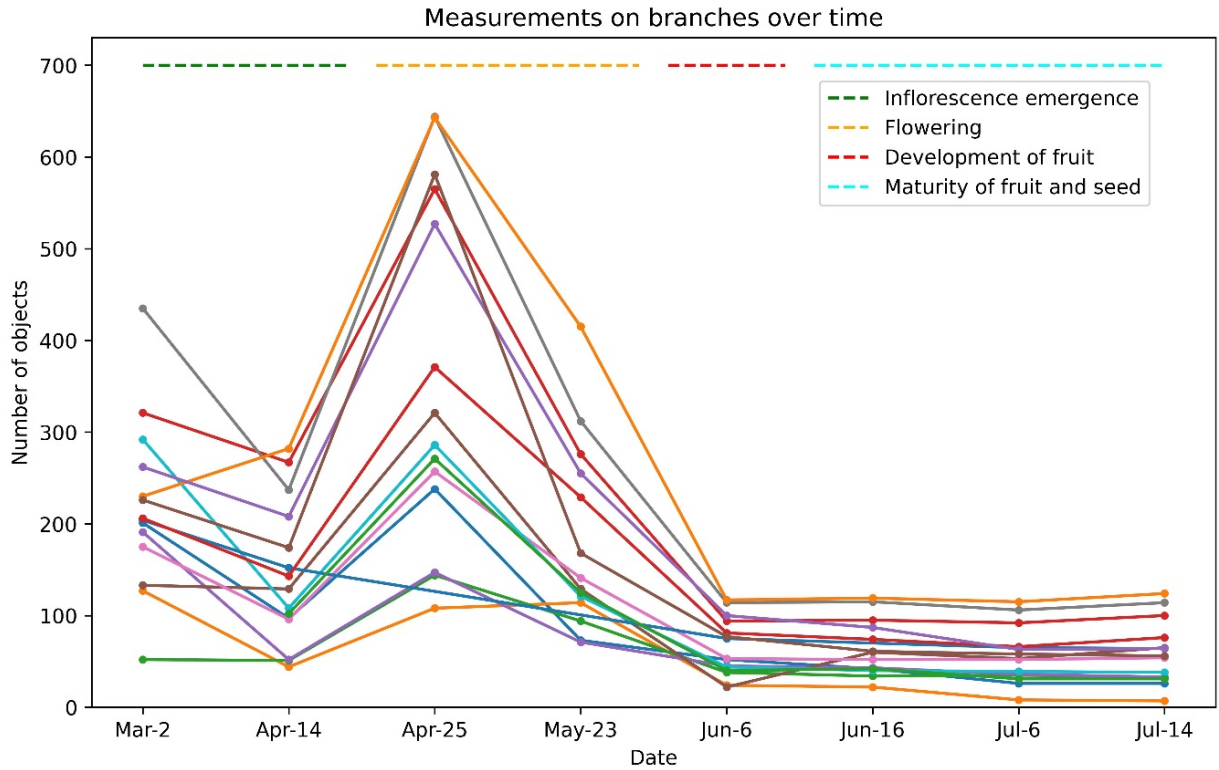

Fig 6. Manually counted references values for every branch during the 2023 season.

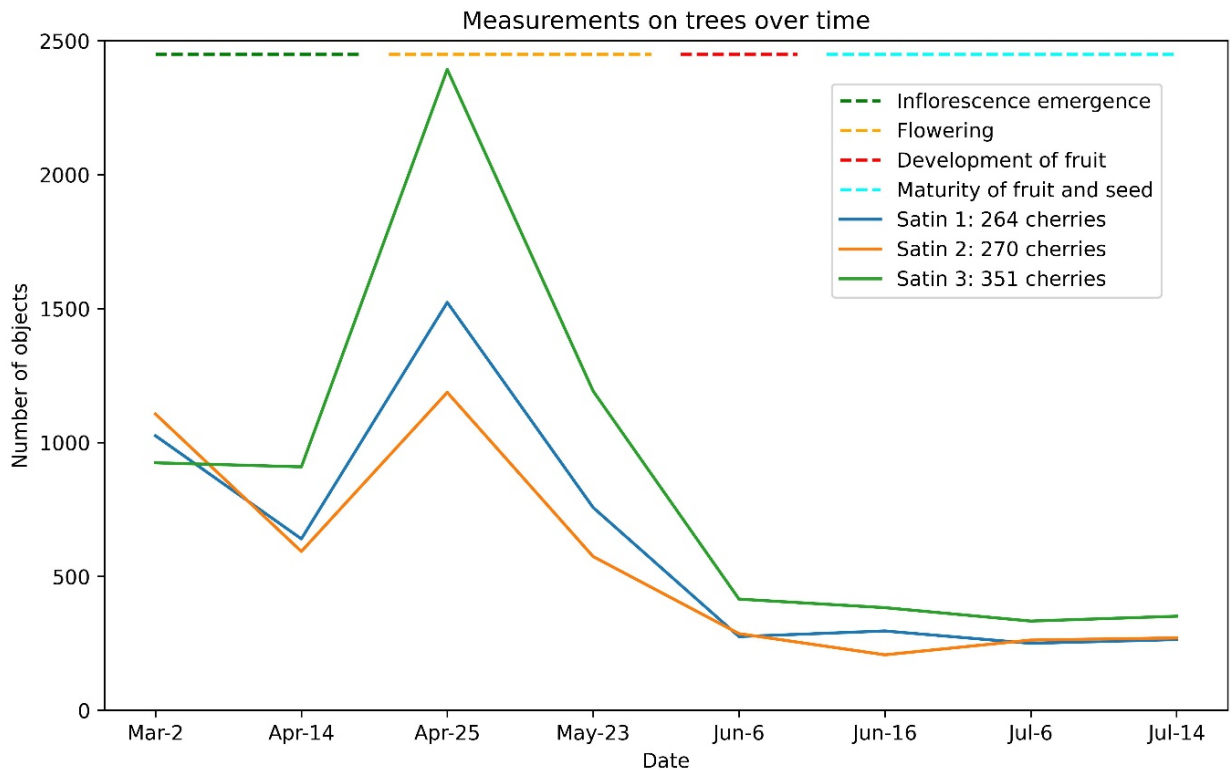

Fig 7. Reference branch values aggregated and grouped by corresponding tree.

**Fruit counting-based yield prediction**

Counting fruit after the second fruit drop (BBCH 73+) could be a good estimator for the final yield. Especially since the risks of natural events with high impact on the fruit count decrease with remaining time until harvest. Because earlier yield estimations are still more valuable than predictions directly before harvest, the time point after the second fruit drop could be a good candidate. As the reference data shows, the fruit count remains relatively stable until harvest.

However, the attempt to count young cherries comes with its own set of challenges. The already introduced *FruitNeRF* framework (Meyer et al., 2024) produced state-of-the-art results for apples, mangoes, plums, and other fruits. Yet it struggled with counting cherries. As seen in Fig. 8, later fruit stages are easy to segment, since sweet cherries have distinctive color and can be detected using simple thresholding procedures. Detection in earlier development stages, is more complicated, but also solvable using modern detection methods, such as supervised deep learning algorithms.

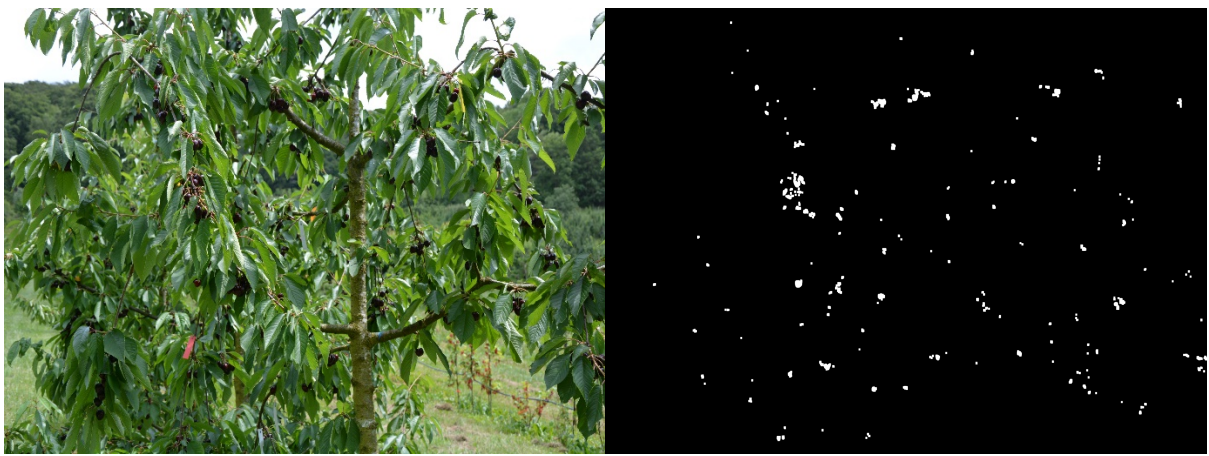

**Fig. 8** Picture of sweet cherry tree with fruit in late development stages (left) together with semantic segmentation mask of fruit based on color thresholding (right).

The cause of the problem is not the 2D segmentation of cherries, but the poor visibility of fruit in images regardless of the viewpoint. This is caused due to inherent traits of cherry trees, like comparatively small fruit size and the fact that cherries tend to grow in clusters that often are poorly visible from multiple angles and tend to stay hidden behind leaves. In photographs of the tree, even from close distance, only a small percentage of the present cherries is visible at all. Basis for the fruit-counting algorithm are semantic 2D segmentation masks of the objects that are to be counted. Fruits that are not present in the image data can thus not be counted later. Consequently, it was not possible to determine a valid fruit count for cherries using *FruitNeRF*. As a future research direction, it could be interesting to find out whether the number of cherries can be estimated based on other tree features instead of accurate counting. To further investigate the informative value of earlier stages in the fruit development cycle, a linear regression analysis was carried out.

**Linear regression for yield prediction**

For accurate yield prediction, there is a strong correlation between the investigated tree trait and the final yield required. Fig. 9 displays the results of the results of a linear regression analysis, which is further underlined by the results shown in Table 2. A significant correlation ($R^2=0.39$, $p<.05$) between the counted objects and harvested cherries is seen as early as during the bud swelling stage on the 2$^{nd}$ of March, more than four months before the harvest. This correlation increases, the closer objects are counted towards the harvest, with the highest correlation less than two weeks before the harvest on the 6$^{th}$ of July ($R^2=0.99$, $p<.001$). Despite these high correlations we observed that the slope of the regression changes considerably throughout the season depending on the type of objects counted. On our first day of measurements on the 2$^{nd}$ of March the slope is 0.23, while counting cherries towards the end of the season resulted almost in one-to-one relationships with the final harvest.

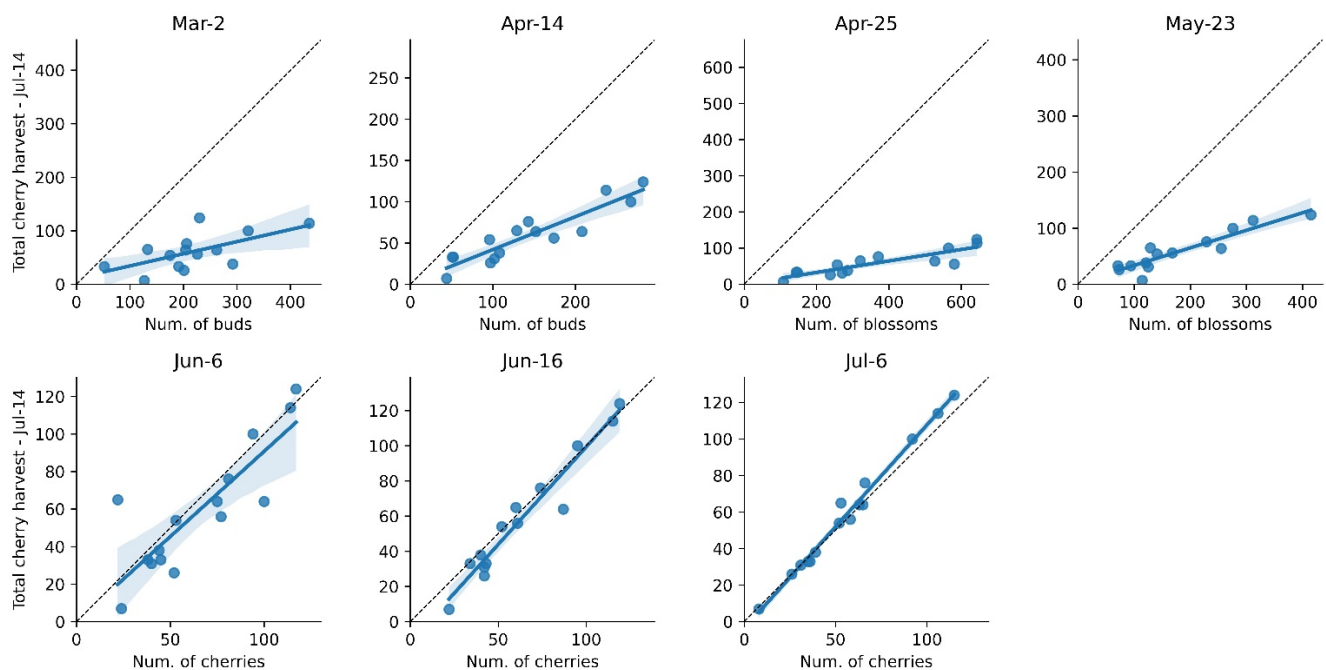

**Fig 9.  Linear regression of objects counted on seven days throughout the season with the final number of cherries counted after harvest on the 14th of July. Each plot shows results for a single day and each measurement point corresponds to the number of objects on an individual branch.**

Table 2.  Results of linear regression analysis. Each row corresponds to a graph in Fig. 9.

| Date | Object | Development stage | BBCH | Slope | Intercept | $R^2$ | p-value | n |
|---|---|---|---|---|---|---|---|---|
| Mar-2 | Buds | Swelling | 51 | 0.23 | 11.49 | 0.39 | P<.05 | 14 |
| Apr-14 | Buds | Open cluster | 56 | 0.40 | 2.03 | 0.84 | P<.001 | 15 |
| Apr-25 | Blossoms | First bloom | 60 | 0.16 | 0.81 | 0.76 | P<.001 | 14 |
| May-23 | Blossoms | Full bloom | 65 | 0.31 | 2.33 | 0.85 | P<.001 | 14 |
| Jun-6 | Cherries | Development of fruit | 75 | 0.91 | -0.10 | 0.72 | P<.001 | 15 |
| Jun-16 | Cherries | Beginning of fruit coloring | 81 | 1.11 | -11.52 | 0.94 | P<.001 | 14 |
| Jul-6 | Cherries | Advanced fruit coloring | 85 | 1.11 | -3.75 | 0.99 | P<.001 | 15 |

The evaluation of the reference data suggests that yield prediction can happen in early vegetation states. Besides the correlation there are several other advantages to consider: Automated evaluation based on image data is easier for trees in vegetative states with no leaves, since leaves occlude the view on developing fruit and cause noisy data and distortions by reacting to natural influences like wind. Earlier yield predictions are also of increasing interest to farmers and breeders since they have more time to adjust their operative and financial planning for the ongoing season. The possible downside of early yield prediction is the increased risk of natural influences or external factors that interfere with the fruit development. While yield predictions during the stage of bud bursting can provide promising results in regular seasons, the risk of a night frost during the flowering stage or a drought later in the season is still present. As part of the For5G project, it was tried to use the results of the 2023 season from Table 2 to predict the yield for the 2024 season. However, there were two frost nights during the flowering stage at the project's field site, which caused an almost complete loss of fruit. While the 2024 harvest is yet to come, it is already clear that the frost damage prevented the application of the 2023 results for the ongoing season.

## Limitations

While the collection of ground truth data was done by domain experts, manual counting still comes with inaccuracies and potential mistakes that should be considered. Occasionally, reference branches were missed or accidentally pruned and during the evaluation of the data some inconsistencies were noticed. Error prune manual counting is however the industry standard and remains the closest approximation to the actual ground truth that is available. Furthermore, the generalizability of the findings is limited because of the small sample size of trees and reference branches. Obtaining data from one field site, during one season is not an ideal approach for findings beyond a proof of concept. The project aims to mitigate this problem by continuing for more seasons and optimizing the data collection strategy for multiple tree rows.

Seasonal changes and weather events can have significant impact on fruit development and the expected yield. A robust yield prediction needs to take this into account, which is difficult because events like night frost, which severely impacted the 2024 season, can only be predicted short term and with limited certainty.

Relying on linear regression as yield prediction has other limitations such as a certain amount of consistency between the trees is required to get accurate factors for estimations. Also, it relies on the assumption that correlation results and factors are transferrable over multiple seasons, which can be wrong, especially for younger trees or species with unregular cycles and higher inter-seasonal variations. Furthermore, good yield prediction relies on accurate feature detection that proves to be challenging with automated techniques, but also with manual counting.

## Conclusion

This paper presented a methodology for obtaining ground truth data of sweet cherries at exemplary time points during an entire season of fruit development. The presented approach has been applied to three *Satin* trees and the results are discussed and visualized.

Regarding the research question of how and when this kind of data is sufficient to support robust yield predictions there are several points that should be considered. Firstly, it can be stated that the final harvested yield can be predicted based on counting in early development stages. Using linear regression gives satisfactory results supporting the hypothesis that the final yield can be predicted reliably. There is a trade-off between benefits and risks of the ideal time for yield predictions. Earlier predictions enable long-term planning and might surpass the prediction skills of a domain expert with years of experience. However, the bigger the time window until harvest, the more unforeseen factors might come into play, which potentially can influence the number of fruits significantly, which makes the early prediction results unusable. Taking this trade-off into account, the evaluation presented in this paper suggests two possible timepoints that seem to be particularly suitable for sweet cherry yield predictions:

For early prediction with higher uncertainty the phase after the buds burst open and clusters emerge from the buds (BBCH 56) seems promising. A less fragile prediction can be made based on counting young cherries after the second fruit drop (BBCH 73+).

Future work should focus on validating these findings on a broader scale as well as developing methods for automated tree feature detection and counting of objects of interest. Various challenges complicate automated data collection and need to be solved for each development stage respectively. Previous unsuccessful experiments with a state-of-the-art fruit-counting algorithm demonstrated that sweet cherry trees are especially challenging for automated feature detection. Considering the impact of external factors such as frost, drought, and birds, which can skew prediction values and assuming consistent tree behavior, it can be concluded that yield prediction with linear progression is possible for all growth stages during the fruit development cycle of sweet cherry trees.

**Acknowledgments**

This project is funded by the 5G innovation program of the German Federal Ministry for Digital and Transport under the funding code 165GU103B.